\documentclass[conference]{IEEEtran}
\IEEEoverridecommandlockouts
\usepackage{cite}
\usepackage{amsmath,amssymb,amsfonts}
\usepackage{algorithmic}
\usepackage{graphicx}
\usepackage{textcomp}
\usepackage{xcolor}
\def\BibTeX{{\rm B\kern-.05em{\sc i\kern-.025em b}\kern-.08em
    T\kern-.1667em\lower.7ex\hbox{E}\kern-.125emX}}
\usepackage{booktabs}
\usepackage{subcaption}
\usepackage{caption}
\usepackage{amsmath}

\usepackage{amssymb}
\usepackage{xspace}
\usepackage{bm}
\usepackage{makecell}
\usepackage{enumitem}
\usepackage{amssymb,mathrsfs,amsmath}

\usepackage{amsmath}

\newcommand{\eg}{\emph{e.g.,}\xspace}
\newcommand{\ie}{\emph{i.e.,}\xspace}
\newcommand{\wrt}{\emph{w.r.t.}\xspace}

\newcommand{\model}{Gaia}

\begin{document}

\title{Gaia: Graph Neural Network with Temporal Shift aware Attention for Gross Merchandise Value Forecast in E-commerce
}
\author{ 
	Borui Ye, Shuo Yang, Binbin Hu, Zhiqiang Zhang, Youqiang He, Kai Huang, Jun Zhou*\thanks{*Corresponding Author}, Yanming Fang \\
	\IEEEauthorblockA{\{borui.ybr, kexi.ys, bin.hbb, lingyao.zzq, heye.hyq, kevin.hk, jun.zhoujun,yanming.fym\}@antfin.com} 
	\textit{Ant Group, Hangzhou, China}
}
\maketitle

\begin{abstract}

E-commerce has gone a long way in empowering merchants through the internet.
In order to store the goods efficiently and arrange the marketing resource properly, it is important for them to make the accurate gross merchandise value (GMV) prediction. However, it's nontrivial to make accurate prediction with the deficiency of digitized data.
In this article, we present a solution to better forecast GMV inside Alipay app. 
Thanks to graph neural networks (GNN) which has great ability to correlate different entities to enrich information, we propose Gaia, a graph neural network (GNN) model with temporal shift aware attention.
Gaia leverages the relevant e-seller' sales information and learn neighbor correlation based on temporal dependencies.
By testing on Alipay's real dataset and comparing with other baselines, Gaia has shown the best performance. 
And Gaia is deployed in the simulated online environment, which also achieves great improvement compared with baselines.
\end{abstract}


\section{Introduction}

With the ever increasing use of the internet, a wide range of small and large companies have leveraged e-commerce to  bolster sales.
Aiming at the sale estimation for each merchant, the forecasting of gross merchandise value (GMV), which estimates the total sales volume over a period of time,
has been playing an increasingly important role in e-commerce scenarios~\cite{yu2021graph,chen2019much}. In addition, a precise prediction has the potential of refraining from unexpected issues for profit loss, \eg stockout, staff inefficiency and customer loss.

\begin{figure}[htp]  
	\begin{minipage}{0.22\textwidth}  
		\centerline{\includegraphics[width=1\textwidth]{"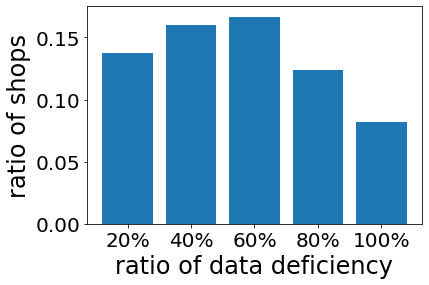"}}
		\centerline{(a)}
	\end{minipage}
	\hfill
	\begin{minipage}{0.20\textwidth}
		\centerline{\includegraphics[width=1\textwidth]{"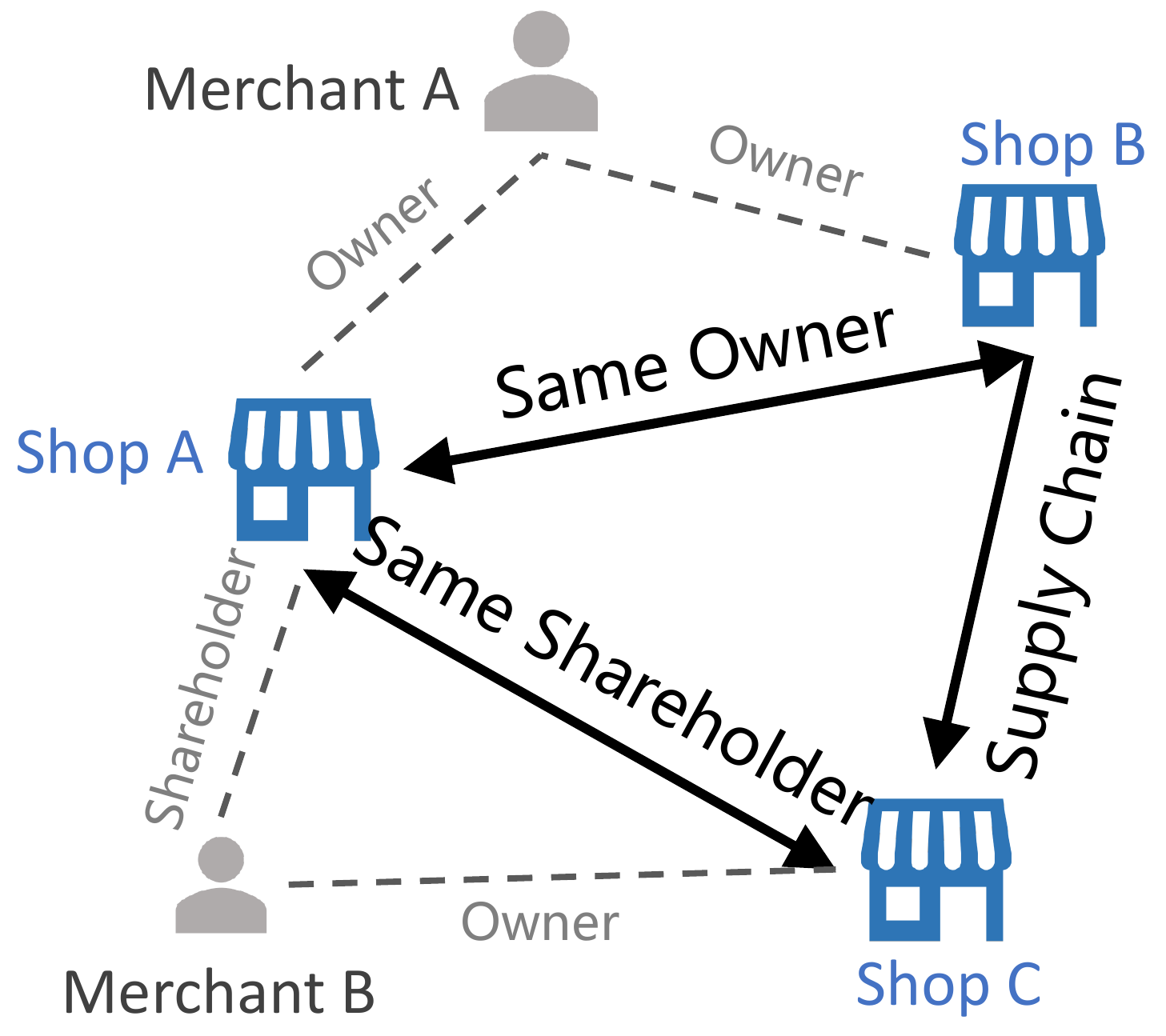"}}
		\centerline{(b)}
	\end{minipage}
	\caption{(a) illustrates the temporal deficiency problem, (b) shows a toy example of e-seller graph, only the blue nodes and the solid line edges exist in the final graph.}
	\label{fig:intro}
\end{figure}

Intuitively, the GMV forecasting task could be formulated as a regression problem with time series analysis, which have been widely explored in numerous studies, including classical statistical method (\eg AR~\cite{box2015time} and ARIMA~\cite{box2015time}) and recently emerging deep learning based methods (\eg LSTM~\cite{hochreiter1997long} and LSTNet~\cite{lai2018modeling}). Unfortunately, the capability of these approaches in time series forecasting may be distant from optimal or even satisfactory, due to the inevitable \textbf{temporal deficiency} issue in practical e-commerce scenarios. Empirically, the skew distribution shown in Fig~\ref{fig:intro}(a) gives the strong evidence that only limited temporal information of e-sellers could be obtained for GMV forecasting, which severely hinders the performance of conventional time series forecasting method.


Besides, the GMV of e-sellers relies heavily on their supply chain enterprises~\cite{yangfinancial}.
This makes graph neural networks~(GNNs) suitable for GMV forcasting for its ability of both utilizing neighbor information and time series dependencies.
In addition, we observe another characteristic of GMV series in the e-commerce scenario, which we call  \textbf{temporal shift}.
We find two kinds of time shifts in GMV series. One is the self-shift, meaning that the GMV series may show similar patterns after an interval. For example, the GMV of a seller who sells seasonal goods may be similar as its historical GMV in the same season~\cite{yu2021graph}.
Another is inter-seller shift, \textit{e.g.} for supply-chain relationships, the GMV of a seller will emerge a rising or decreasing pattern earlier than its downstream retailers, as retailers will firstly buy goods from its suppliers then retail to the customers.
Though some GNN-based methods~\cite{yu2018spatio, wu2019graph, zheng2020gman} have been proposed for GMV forcasting, they fail to make full use of time shift information in the e-commerce scenario.

We hereby propose Gaia, a \underline{G}r\underline{a}ph Neural Network with Tempoal Sh\underline{i}ft aware \underline{A}ttention. 
To solve the \textbf{temporal deficiency} problem, we mine different relationships among shops and establish a shop network, making up for the information loss.
Our model contains three basic components: \underline{F}eature \underline{F}usion \underline{L}ayer~(FFL), \underline{T}emporal \underline{E}mbedding \underline{L}ayer (TEL) and an \underline{I}nter and intra \underline{T}emporal shift \underline{A}ware Attention mechanism for classical \underline{G}raph \underline{C}onvolutional \underline{N}etwork (short as ITA-GCN).
The FFL is in charge of combining basic features and GMV series of a shop, TEL aims at extracting temporal patterns, and the ITA-GCN is well-designed to caputure the \textbf{Temporal Shift} patterns not only from the time series of an individual shop, but also from its neighbors.

Our main contribution is listed as follows:
\begin{itemize}
	\item In the e-commerce scenario, we propose Gaia, a GMV forecasting framework, which aims at solving the \textbf{temporal deficiency} problem and the \textbf{temporal shift} problem of e-sellers.
	\item Our proposal not only fuses auxiliary features with time series features of a e-seller via a FFL and a TEL, but also learns temporal shift patterns from its own and its neighbors' GMV sequence, via a ITA-GCN.
	\item We conduct extensive experiments on real-world data set and also deploy Gaia in online environment, both the offline and online results show that Gaia outperforms all state of the art methods.
\end{itemize}

\section{Related Work}
In past decades, time series analysis based methods have been well studied~\cite{aminikhanghahi2017survey}.
Due to the simplicity and interpretability, early works mainly focus on statistical modelling based on a assumption of stationary process~\cite{box2015time}. Particularly, the ARIMA approach and its variants (\eg AR, MA, and ARMA) have shown powerful capability in various applications~\cite{box2015time}, which make prediction only by the linear combination of historical values, \ie so-called  univariate time series analysis. On the comparison, attention is natually shifting towards multivariate time series analysis and a series of approaches~\cite{durichen2014multitask} are proposed to inherently characterize interdependencies among variables. However, the prefabricated assumption and model complexity are bottlenecks. 
Due to the powerful ability of feature interaction, deep neural networks are introduced to capture non-linear patterns in time series. Following this line, DeepAR~\cite{salinas2020deepar} and LogTrans~\cite{li2019enhancing} are respectively developed for univariate time series prediction based on deep  probabilistic models and the recently emerging Transformer architecture. Meanwhile, to marry the strength of convolutional neural networks (CNNs) and recurrent neural networks (RNNs), numerous methods~\cite{lai2018modeling,maggiolo2019autoregressive,shih2019temporal} are proposed to capture local and global dependencies among variables for  multivariate time series prediction. Unfortunately,  the  inevitable temporal  deficiency issue in practical e-commerce scenarios still threatens the capbility of current methods.


As a prevailing paradigm, graph neural networks (GNNs) has shown the remarkable strength for ingesting  valuable information encoded in graph-structured data~\cite{wu2020comprehensive,zhang2020deep,velivckovic2017graph,hamilton2017inductive,kipf2016semi}. In line with the main focus in our paper, we center on the well studied spatial-temporal GNNs (STGNNs) which initially designed to the traffic prediction task~\cite{yu2018spatio,zheng2020gman,chen2020multi}. In general, a STGNN is comprised of two main components: a graph convolution capturing spatial structure and a deep architecture dealing with time series on nodes through CNNs~\cite{yan2018spatial,yu2018spatio} and RNNs\cite{li2017diffusion,seo2018structured,xie2020deep}. Distinct from above paradigm based on pre-defined graph structure, a few efforts~\cite{wu2019graph,wu2020connecting} have been made for simultaneously learning a graph structure and a powerful GNN for time series prediction in an unified framework.
Nevertheless, these methods still neglect the temporal shift issues. In contrast, our proposed {\model} hinges on a well-designed graph convolutional component, which carefully considers the temporal shift in both inter and intra level. In addition, node-level feature learning is also achieved in a more fine-grained manner through feature fusion and temporal embedding.

\section{Preliminaries}
\subsection{Problem Definition}
This paper addresses the problem of forecasting GMV of e-sellers. Suppose there exists an e-seller graph $\mathcal{G}=(\mathcal{V}, \mathcal{E})$  consisting of $N$ e-sellers as node set $\mathcal{V}=\{v_1, \ldots, v_N\}$, and $M$ links between these e-sellers as edge set $\mathcal{E}=\{e_1, \ldots, e_M\}$. 
For each e-seller $v$, let $\bm{z}_v \in \mathbb{R}_{+}^{T}$ denote its monthly GMV, where $T$ is the number of time steps.
We assume that each e-seller have both temporal and static auxiliary features (detail description in Section \ref{sec:ffm}), which can be denoted as $\mathbf{F}^\mathcal{T}_v = \{\bm{f}^\mathcal{T}_{v,t}\}_{t=1}^T \in \mathbb{R}^{T \times D^{\mathcal{T}}}$ and $\bm{f}^\mathcal{S}_v \in \mathbb{R}^{D^{\mathcal{S}}}$, respectively.
Given the above information, the problem is to predict the future GMV of $T^{\prime}$ months for each e-seller $v$, which can be denoted as $\bm{y}_v \in \mathbb{R}_{+}^{T^{\prime}}$.
\subsection{E-Seller Graph Construction}
In the scenario of e-commerce, there exists two kinds of relationships between e-sellers that can improve the effectiveness of GMV forecasting. 
Fig~\ref{fig:intro}(b) shows a demonstration of these two relationships. 
The first relationship is \textbf{supply chain relationship}, in which one associated e-seller sells its goods to another e-seller, and the upstream e-seller's GMV is affected by the downstream e-seller.
The GMV traded by adjacent e-sellers in a supply chain is usually correlated. For example, a downstream e-seller with an increased GMV may need more raw materials to produce more goods. This may lead to an increase in orders from its upstream suppliers, thus increase the GMV of upstream e-sellers to a certain extent.
The second one is \textbf{same owner/shareholder relationship}. Since two e-sellers that share the same owners or shareholders usually have similar operation strategies, such as similar willingness to participate in shopping festivals, their GMVs may share a common trend. 

These two kinds of relationships make up the edge set of e-seller graph. Note that, the e-seller consists of only shops as nodes, and the edges type is made as one of the edge features, so the graph here is considered a homogenous graph.
\section{Methodology}


\begin{figure*}
	\centering
	\includegraphics[width=0.95\textwidth]{"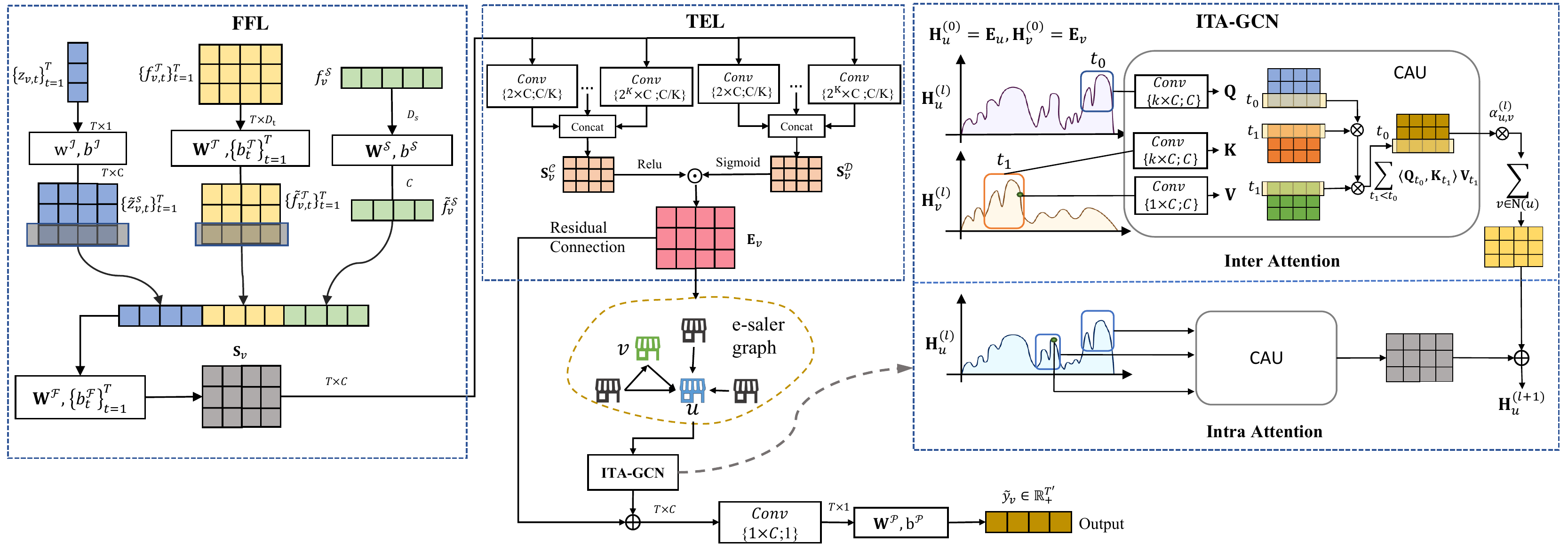"}
	\caption{Overview of {\model}, which consists of three main components: \underline{F}eature \underline{F}usion \underline{L}ayer (FFL), \underline{T}emporal \underline{E}mbedding \underline{L}ayer (TEL) and \underline{I}nter and intra \underline{T}emporal shift aware \underline{A}ttention based \underline{G}raph \underline{C}onvolutional \underline{N}etwork (ITA-GCN). FFL first fuses auxiliary features with the GMV series at a single timestamp.
    Then TEL models the temporal patterns along the timeline.
    Finally, with an well-designed convolutional attention unit, ITA-GCN learns the temporal shift on the structured e-seller graph.}
	\label{fig:ov}
\end{figure*}

The overview of {\model} is depicted in Fig. \ref{fig:ov}.
\subsection{Feature Fusion Layer}
\label{sec:ffm}
In the industrial scenario of e-commerce, an e-seller is usually characterized via various features. Here, for each e-seller node $v$, we summarize following features:

\begin{itemize}
    \item $\{z_{v,t}\}^{T}_{t = 1}$:  historical monthly GMV series.
    \item $\bm{f}^{\mathcal{T}}_{v,t} \in \mathbb{R}^{D^{\mathcal{T}}}$: auxiliary temporal features at each time $t \in [1, 2, \cdots, T]$, \eg the month, the monthly amount of customers and orders.
    \item $\bm{f}^{\mathcal{S}}_v \in \mathbb{R}^{D^{\mathcal{S}}}$:  auxiliary static features, \eg the industry, the registration location.
\end{itemize}
Intuitively, both the individual time series and auxiliary temporal/static features could help {\model} better understand the intrinsic change trend of GMV for each e-seller.
For better representation, we equip {\model} with a \emph{\underline{F}eature \underline{F}usion \underline{L}ayer} to fuse available features in a more fine-grained manner.
Formally, given an e-seller $v$ at time {t}, the FFL firstly projects aforementioned features (\ie $z_{v,t}, \bm{f}^{\mathcal{T}}_{v,t}$ and $\bm{f}^{\mathcal{S}}_v$ ) to the $C$-dimensional embedding space, followed by a concatenation and a fully-connected layer for feature fusion.
\begin{align}
\small
\tilde{\bm{z}}_{v,t} & = z_{v,t} \cdot \bm{w}^{\mathcal{I}} + \bm{b}^{\mathcal{I}}  ,\\
\tilde{\bm{f}}^{\mathcal{T}}_{v,t}&=\mathbf{W}^{\mathcal{T}} \bm{f}_{v,t}^{\mathcal{T}}+ \bm{b}^{\mathcal{T}}_{t} , \\
\tilde{\bm{f}}_v^{\mathcal{S}}&=\mathbf{W}^{\mathcal{S}} \bm{f}_v^{\mathcal{S}}+ \bm{b}^{\mathcal{S}} ,\\
\bm{s}_{v,t} &=\mathbf{W}^{\mathcal{F}}[\tilde{\bm{z}}_{v,t}\ ||\ \tilde{\bm{f}}_{v,t}^{\mathcal{T}}\ ||\ \tilde{\bm{f}}_{v}^{\mathcal{S}}]+ \bm{b}^{\mathcal{F}}_t,
\end{align}
where $\bm{w}^{\mathcal{I}} \in \mathbb{R}^{C}$, $\bm{b}^{\mathcal{I}} \in \mathbb{R}^C$, $\mathbf{W}^{\mathcal{T}} \in \mathbb{R}^{C \times D^{\mathcal{T}}}$, $\{\bm{b}^{\mathcal{T}}_{t}\in \mathbb{R}^{C}\}_{t = 1}^{T}$, $\mathbf{W}^{\mathcal{S}} \in \mathbb{R}^{C \times D^{\mathcal{S}}}$, $\bm{b}^{\mathcal{S}}\in \mathbb{R}^C$, $\mathbf{W}^{\mathcal{F}} \in \mathbb{R}^{C\times 3C}$ and $\{\bm{b}^{\mathcal{F}}_{t}\in \mathbb{R}^{C}\}_{t = 1}^{T}$ are learnable parameters, ``$||$'' 
is the concatenation operator.

\subsection{Temporal Embedding Layer}
As mentioned above, the FFL endows {\model} with powerful capability for subtle feature fusion at each single timestamp, whereas the feature interaction along the timeline is not carefully considered, which potentially implies the various temporal patterns (\eg annual and seasonal patterns) derived from GMV series.  Therefore, inspired by the idea of ~\cite{wu2020connecting}, we develop the  \emph{\underline{T}emporal \underline{E}mbedding \underline{L}ayer} (TEL) based on coupled temporal convolution layers, where a temporal convolution layer is in charge of temporal patterns extraction, paired with another temporal convolution layer for information denoising.
Specifically, given an input GMV series for e-seller $v$ (\ie $\{z_{v,t}\}^{T}_{t = 1}$), we could obtain the fused features at each time step through the FFL, denoted as  temporal feature matrix $\mathbf{S}_v = \{ \bm{s}_{v,t} \}_{t = 1}^{T} \in \mathbb{R}^{T \times C}$. 
Following common strategies in \cite{fisher2016iclr,liu2020multi}, we define a group of kernels with size $\{(2^{k} \times C ; C / {K})\}_{k = 1}^{K}$, which aim at capturing temporal patterns in multiple levels. Here, $\{a \times b; c\}$ means a subgroup of $c$ kernels with size $\{a \times b\}$.
Then, the coupled temporal convolution layers work as follows:
\begin{eqnarray}
\small
\mathbf{S}_v^{\mathcal{C}} = [\mathbf{L}_{\{2 \times C; C/K\}}^{\mathcal{C}, 1} \star \mathbf{S}_v\  ||\ \cdots\  ||\ \mathbf{L}_{\{2^K \times C; C/K\}}^{\mathcal{C}, K} \star \mathbf{S}_v],\\
\mathbf{S}_v^{\mathcal{D}} = [\mathbf{L}_{\{2 \times C; C/K\}}^{\mathcal{D}, 1} \star \mathbf{S}_v\  ||\ \cdots \  ||\ \mathbf{L}_{\{2^K \times C; C/K\}}^{\mathcal{D}, K} \star \mathbf{S}_v],
\end{eqnarray}
where $\mathbf{L}^{\mathcal{C}, k}_{\{a \times b;c\}}$ and  $\mathbf{L}^{\mathcal{D}, k}_{\{a \times b;c\}}$ means the $k$\textsuperscript{th} temporal convolution consisting of $c$ kernels with size $\{a \times b\}$ for temporal pattern capturing and information denoising, respectively, and $\star$ is the 1D convolution operator with zeros padding.

Clearly, above temporal convolution layers provide coupled valuable matrices, where the former (\ie $\mathbf{S}_v^{\mathcal{C}} \in  \mathbb{R}^{T \times C}$ ) preserves multi-level temporal patterns derived from GMV series while the latter (\ie $\mathbf{S}_v^{\mathcal{D}} \in  \mathbb{R}^{T \times C}$ ) emphasizes important/relevant patterns. Subsequently, the TEL gives the final temporal representation for e-seller $v$ as follows:
\begin{equation}
\mathbf{E}_v=\text{ReLU}(\mathbf{S}_v^{\mathcal{C}}) \odot \text{Sigmoid}(\mathbf{S}_v^{\mathcal{D}}),
\end{equation}
where $\odot$ denotes the Hadamard product.


\subsection{Inter and  Intra Temporal  shift Aware Attention based Graph Neural Network}
In this section, we are devoted to learning structural information from the well-established e-seller graph to benefit GMV forecasting.   
As mentioned above, we should pay careful attention to following two temporal shift issues in e-commerce scenarios, which could be hardly captured by current graph neural networks:

\begin{itemize}[left=0pt,nosep]
\item \textbf{Inter temporal shift} among two connected e-sellers. Generally, different types of e-seller (\ie suppliers and retailers) have different response times to market trends, \eg the GMV of suppliers usually increase/decrease several months before retailers. 
Moreover, a center e-seller in the graph may place different importances to its neighbors, where neighbors with rich series are expected to be emphasized.
\item \textbf{Intra temporal shift} existed in individual e-seller. Intuitively, the GMV of a certain e-seller is directed related to time and varies with seasons, \ie the GMV of a e-seller selling  air conditioners always a sharp rise in summer. 
Such a periodic shift is essential to accurately summarize historical GMV series for each individual shop.
\end{itemize}
In light of these findings, we prepare a well-designed \underline{I}nter and intra \underline{T}emporal shift aware \underline{A}ttention mechanism for classical \underline{G}raph \underline{C}onvolutional \underline{N}etwork (short as ITA-GCN) to  tackle the above issues.

\subsubsection{Convolutional Attention Unit}
As the heart of ITA mechanism, \underline{C}onvolutional \underline{A}ttention \underline{U}nit (short as CAU), based on recently emerging self-attention architecture~\cite{vaswani2017attention}, aims at capturing temporal shift for arbitrary edge $v \rightarrow u$ ($u$ and $v$ could be the same). In other words, the CAU learns temporal attention weights over timestamps conditioned on paired GMV series. Given an edge $v \rightarrow u$, the CAU produce the final representation that summarize the influence of temporal shift from $v$ to $u$ as follows: 
\begin{eqnarray*}
\small
\mathbf{Q}_u&=&\mathbf{L}_{\{3 \times C; C\}}^Q \star \mathbf{H}_u, \\
\mathbf{K}_v&=&\mathbf{L}_{\{3 \times C; C\}}^K \star \mathbf{H}_v, \\
\mathbf{V}_v&=&\mathbf{L}_{\{1 \times C; C\}}^V \star \mathbf{H}_v, \\
\mathrm{CAU}(\mathbf{H}_u, \mathbf{H}_v)&=&
	\mathrm{softmax}(\frac{\mathbf{Q}_u\mathbf{K}^\top_v}{\sqrt{C}} + \mathbf{M})\ \mathbf{V}_v
\end{eqnarray*}
where $\mathbf{H}_u, \mathbf{H}_v \in \mathbb{R}^{T\times C}$ is temporal representation for node $u$ and $v$, respectively. It is worthwhile that convolutional kernels (\ie $\mathbf{L}^Q, \mathbf{L}^K, \mathbf{L}^V$) are incorporated  to help CAU be aware of locality~\cite{li2019enhancing} of GMV series so that relevant features based on the shape of several adjacent points could be correctly matched. Moreover, we employ a mask matrix $\mathbf{M} \in \{-\infty,0\}^{T\times T}$ for filtering out rightward attention in order to avoid future information leakage.


\subsubsection{ITA-GCN layer}
Next, we build upon the architecture of graph attention network~\cite{velivckovic2017graph} to recursively obtain center node's representation by aggregating its neighbors on e-seller graphs. Moreover, attentive weights of aggregation are generated to distinguish influence of temporal shift passed by connectivity. Here, we begin with a single layer, which produces representation for center node $\mathbf{H}_u^{(l+1)}$ by capturing 
i) inter temporal shift from influences of its neighbors and 
ii) intra temporal shift from its historical GMV series
through our CAU component.
\begin{equation}
\small
\mathbf{H}_u^{(l + 1)}=
\underbrace{\sum_{v\in N(u)}\alpha_{u,v}^{(l)}
\mathrm{CAU}(\mathbf{H}_u^{(l)}, \mathbf{H}_v^{(l)})}_\text{Inter Neighbor Attention}
+ \underbrace{\mathrm{CAU}(\mathbf{H}_u^{(l)}, \mathbf{H}_u^{(l)})}_\text{Intra Self Attention},
\end{equation}
where $N(u)$ is the neighbor set of node $u$,  $\mathbf{H}_u^{(l)}$ and $\mathbf{H}_v^{(l)}$ is the representation of $l$-th ITA-GCN layer for  node $u$ and $v$ respectively, which is initialized with the output of TEL, \ie $\mathbf{H}_u^{(0)}=\mathbf{E}_u$ and $\mathbf{H}_v^{(0)}=\mathbf{E}_v$. Moreover,  $\alpha_{u,v}^{(l)}$ controls how much information being aggregated on edge ``$u \leftarrow v$'', which is implemented as follows: 
\begin{eqnarray*}
    \small
	\alpha_{u,v}^{(l)} &=&
	\frac{\text{exp}\ g(u, v)}{\sum_{v^{\prime} \in N(u)} \text{exp}\ g(u, v^{\prime})}\\
	g(u,v) &=& \bm{\mu}^\top\tanh(\mathbf{L}^s_{\{1 \times C;1\}} \star \mathbf{H}_u^{(l)} +\mathbf{L}^d_{\{1 \times C;1\}} \star \mathbf{H}_v^{(l)})
	\label{eq:geniepath}
\end{eqnarray*}
Where $\bm{\mu} \in \mathbb{R}^{T}$ is model parameter, $\mathbf{L}^s_{\{1 \times C;1\}}$ and $\mathbf{L}^d_{\{1 \times C;1\}}$ are convolution kernels.

\subsection{Model Learning}
Generally, we stack $L$ ITA-GCN layers to fully capture complicated structure implicated in e-seller graph, and denote the final representation for target node $u$ as $\mathbf{H}_u^{(L)}$. Then, the GMV of node $u$ in future $T'$ months could be predicted as follows:
\begin{equation}
	\mathbf{\tilde{y}}_u = \text{ReLU}([\mathbf{L}_{\{1 \times C; 1\}}^{\mathcal{P}} \star (\mathbf{H}_u^{(L)} + \mathbf{E}_u)] \mathbf{W}^{\mathcal{P}} + \bm{b}^{\mathcal{P}}).
	\label{eq:output}
\end{equation}
Here, our prediction function is parameterized by  weight matrix $\mathbf{W}^{\mathcal{P}} \in \mathbb{R}^{T \times T^\prime}$, bias vector $\bm{b}^{\mathcal{P}} \in \mathbb{R}^{T^\prime}$ as well as convolutional kernel $\mathbf{L}_{\{1 \times C; 1\}}^{\mathcal{P}}$, and we incorporate the  residual connection mechanism to emphasize the original representations derived from TEL model.

Since GMV forecasting could be naturally formulated as a regression task, \underline{M}ean \underline{S}quare \underline{E}rror (MSE) is adopted to guide the optimization of {\model}.
\begin{equation}
	\mathcal{L} = \frac{1}{|\mathcal{V}| \times T^{\prime}} \sum_{u \in \mathcal{V}} \sum_{t=1}^{T^\prime} (\mathbf{\tilde{y}}_{u,t} - \mathbf{y}_{u,t})^2,
	\label{eq:loss}
\end{equation}
where $\mathbf{\tilde{y}}_{u,t}$ and $\mathbf{y}_{u,t} \in \mathbb{R}_{+}^{T^\prime}$ is the prediction of {\model} and ground truth in $t$-th month.

\section{Experiment}

\subsection{Experimental Setup}

\begin{table*}
	\caption{Performance comparison with baselines on three datasets}
	\label{tab:baseline}
\centering
\setlength{\tabcolsep}{3.5mm}{
	\begin{tabular}{c|ccc|ccc|ccc}
		\toprule
      	&& Oct. & & & Nov. &&& Dec. &\\
		Method&MAE $\downarrow$ &RMSE $\downarrow$ &MAPE $\downarrow$ &MAE $\downarrow$ &RMSE $\downarrow$&MAPE $\downarrow$&MAE $\downarrow$&RMSE $\downarrow$ &MAPE $\downarrow$\\
		\midrule
		ARIMA & 39,493& 139,405 & 0.2145 & 40,329 & 142,378 & 0.2427 & 38,148 & 104,654 &0.2010\\
		LogTrans & 43,337 &550,485 & 0.1293 & 42,895 & 532,192 & 0.1165 & 41,884 & 550,884 & 0.1041\\
		\midrule
		GAT & 42,119 &472,615 &0.1557 & 39,961& 441,983&0.1462 & 37,952 & 452,788&0.1258\\
		GraphSage & 40,195 &503,052 & 0.1386 & 38,417 & 472,788 &0.1314& 37,278 & 482,840&0.1168\\
		Geniepath & 40,472 & 480,509 & 0.1475 & 38,543 & 457,190 & 0.1380 & 36,753 & 466,391 & 0.1189\\
		\midrule
		STGCN & 42,413 &544,015& 0.1389 &39,099&514,525&0.1261& 36,368&522,495&0.1042\\
		GMAN & 39,889 & 412,678 & 0.1391 & 37,467 &400,293&0.1298& 34,240 & 402,699&0.1101\\
		MTGNN & 28,721 &158,596  & 0.1089 & 26,346 & 141,067& 0.0992 & 24,357 & 167,072& 0.0871\\
		\midrule
		Gaia & \textbf{24,064} & \textbf{112,516} & \textbf{0.0909} & \textbf{22,467} & \textbf{95,518} & \textbf{0.0860} & \textbf{20,473} & \textbf{95,051} & \textbf{0.0771}  \\
		\bottomrule
	\end{tabular}}
\end{table*}


\subsubsection{Evaluation Dataset and Metrics}
We conduct experiments on real-world datasets from Alipay, which contain 3 million of shops over the time period from Jun. 2019 to Dec. 2020. To evaluate the performance of each method, we utilize the data from Jun. 2019 to Sep. 2020 and perform GMV forecasting for shops in the remaining three months (\ie \textbf{Oct.}, \textbf{Nov.} and \textbf{Dec.}). The dataset follows this data statement:
\begin{enumerate}
\item It does not contain any Personal Identifiable Information.
\item It's desensitized and encrypted.
\item Adequate data protection was carried out during the experiment to prevent the risk of data copy leakage, and the dataset was destroyed after the experiment.
\item It's only used for academic research, it does not represent any real business situation.
\end{enumerate}

To guarantee the stability of prediction, we define the label of each shop as its total GMV in the future 3 months.
For each shop, we collect its historical monthly GMV in the last 24 months from online order logs to construct GMV series.
Moreover, we carefully construct an e-seller graph to help GMV forecasting, which consists of around 3 million of nodes (\ie shops) and 10 million of edges (\ie same owner/shareholder relationship and supply chain relationship). And the supply chain relationship is mined as introduced in~\cite{yang2021inductive,yangfinancial}.


Following~\cite{bergmeir2018note}, we adopt widely-used \textbf{MAPE}, \textbf{RMSE}, \textbf{MAE} to evaluate performance on the GMV forecasting task.



\subsubsection{Compared Methods}
We mainly consider 9 representative methods for the GMV forecasting task, which falls into three groups: 
i) Time series analysis based methods (\ie \textbf{ARIMA}~\cite{box2015time} and \textbf{LogTrans}~\cite{li2019enhancing}) only utilizing the individual sequential data,  
ii) GNN based methods (\ie \textbf{GAT}~\cite{velivckovic2018graph}, \textbf{GraphSAGE}~\cite{hamilton2017inductive} and \textbf{GeniePath}~\cite{liu2019geniepath}) only considering the graph structure,
and iii) STGNN based methods (\ie \textbf{STGCN}~\cite{yu2018spatio}, \textbf{GMAN}~\cite{zheng2020gman} and \textbf{MTGNN}~\cite{wu2020connecting}) jointly modelling sequential and structural information through so-called spatial and temporal attention mechanism.
\subsubsection{Implementation Details}
With AGL~\cite{zhang2020agl} framework, we use Keras, and adopt Adam~\cite{kingma2014adam} optimizer with learning rate $0.00001$ and $32$ batch size. For fair comparison, we set embedding sizes to $32$. To obtain optimal performance of each methods, we apply the grid search strategy on the validation set, and  optimal hyper-parameters used in {\model} and baselines are listed as follows:
For time series analysis based methods, we set the key parameters (\ie max($p$) and max($q$)) in \textbf{ARIMA} to 2; for \textbf{LogTrans} we use $3$ attention blocks with $3$ heads.
For GNN based methods, we follow the same architectures in their original paper, and stack 2 layers for information aggregation.
For STGNN based methods, we set the channel size to 32. Specifically, \textbf{MTGNN}'s layer size is set to 3.


\subsection{Experimental Results and Analysis}

\subsubsection{Overall Comparison}
We present the comparison results of {\model} and compared methods in Table~\ref{tab:baseline}. 
We can observe that our model consistently and significantly outperforms all baselines on three datasets across all metrics, demonstrating the effectiveness of {\model} on the task of GMV forecasting.
Moreover, the overall performance order among baselines are follows: STGNN based methods $>$ GNN based methods and $>$ time series analysis based methods. It is not surprising that GMV forecasting could easily benefit from fusion of auxiliary information (\eg graph structure). 
Nevertheless, our model still yields the best performance by jointly characterizing individual feature interaction and graph based incorporation in a more fine-grained manner.

\subsubsection{Ablation Study}
We conducted a comprehensive ablation study to analyze the impacts of each well-designed component in our architecture. Table~\ref{tab:ab2} shows the performance of {\model} ans its variants on three datasets. The variants and corresponding analysis are listed as follows:
\begin{itemize}
    \item \underline{I}nter and intra \underline{T}emporal shift aware \underline{A}ttention ({\model} w/o ITA): We replace the newly proposed ITA with traditional self-attention. The significant performance drop further supports the conclusion that inter and intra temporal shift should be carefully handled in structural learning with our e-seller graph.
    \item \underline{F}eature \underline{F}usion \underline{L}ayer  ({\model} w/o FFL): Not surprisingly, we observe poor performance without our FFL, implying that feature fusion in a more fine-grained mode plays a fundamental role in following graph learning and final GMV forcasting.
    \item \underline{T}emporal \underline{E}mbedding \underline{L}ayer  ({\model} w/o TEL):  We utilize one certain convolutional kernel (\ie $\{4 \times C; C\}$) rather that kernel group in our TEL. Clearly, the experimental results shows that a single kernel is not enough for various temporal patterns in GMV series.
\end{itemize}

\begin{table}
	\caption{Ablation Study of {\model}}
	\label{tab:ab2}
	\centering
	\setlength{\tabcolsep}{3.0mm}{
	\begin{tabular}{c|l|ccc}
		\toprule
		& \makecell[c]{Method}& MAE $\downarrow$ & RMSE $\downarrow$     & MAPE $\downarrow$ \\
		\midrule
		& Gaia & \textbf{24,064} & 112,516 & \textbf{0.0909}   \\
	{Oct.}	& \ w/o ITA &26,387 & 131,523 &0.0955  \\
		& \ w/o FFL &26,217 & 131,689 & 0.1002 \\
		& \ w/o TEL & 27,021 & 103,771 & 0.1017 \\
		 \midrule
		& Gaia & \textbf{22,467} & \textbf{95,518} & \textbf{0.0860}   \\
	{Nov.}	& \ w/o ITA &24,115  & 131,470 & 0.0876  \\
		& \ w/o FFL &23,915 &141,535 & 0.0910 \\
		& \ w/o TEL & 24,816 & 127,711 &  0.0929 \\
		 \midrule
		& Gaia & \textbf{20,473} & \textbf{95,051} & 0.0771   \\
	{Dec.}	& \ w/o ITA & 21,551 &  153,490 & 0.0767  \\
		& \ w/o FFL &21,305  & 134,152 &0.0791 \\
		& \ w/o TEL & 22,458 & 117,293 & 0.0817\\
		 \bottomrule
	\end{tabular}}
\end{table}


\subsubsection{Effectiveness Analysis of  Graph}
\begin{figure}[htp]  
	\begin{minipage}{0.24\textwidth}  
		\centerline{\includegraphics[width=1\textwidth]{"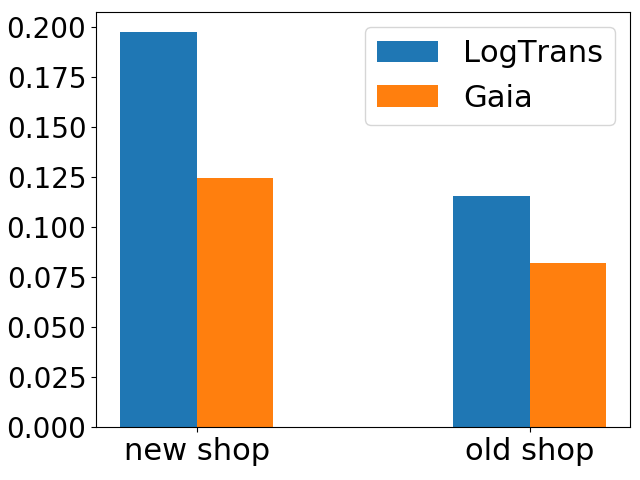"}}
		\centerline{(a) MAPE}
	\end{minipage}
	\hfill
	\begin{minipage}{0.24\textwidth}
		\centerline{\includegraphics[width=1\textwidth]{"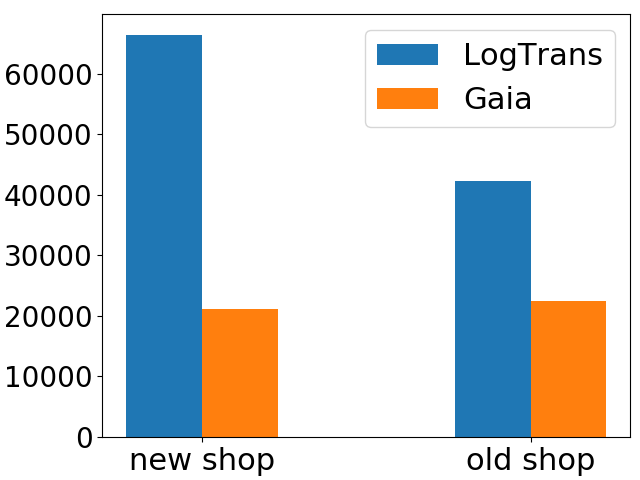"}}
		\centerline{(b) MAE}
	\end{minipage}
	\caption{Effectiveness Analysis of e-seller Graph. Larger  performance  margin  between {\model}  and  LogTrans  on  the  ``New  Shop  Group'' could be observed. }
	\label{fig:lags_pg}
\end{figure}

Next, we take a closer look at the effective of our e-seller Graph towards {\model}, which devoted to the inevitable temporal deficiency issue in practical e-commerce scenarios. According to the length of the GMV series, we categorize all shops in our datasets into two groups, \ie ``New Shop Group'' with $T < 10$ and  ``Old Shop Group'' with $T >= 10$. And we select the strongest baseline without graph for comparison and report the performance \wrt MAPE and MAE in Fig~\ref{fig:lags_pg}. 

From the figures, we could observe that {\model} significantly outperforms LogTrans with the help of graph learning. More importantly, we find the larger performance margin between {\model} and LogTrans on the ``New Shop Group'', \ie 215.8\% \wrt MAE and 58.8\% \wrt MAPE  improvements on ``New Shop Group'' $\bm{v.s.}$ 88.5\% \wrt MAE and 41.0\% \wrt MAPE  improvements on ``Old Shop Group''. Both findings further demonstrate the superior capacity of {\model} for addressing the temporal deficiency issue with e-seller graph.

\subsubsection{Case Study towards the ITA module}
\begin{figure}[htp]  
	\begin{minipage}{0.24\textwidth}  
		\centerline{\includegraphics[width=1\textwidth]{"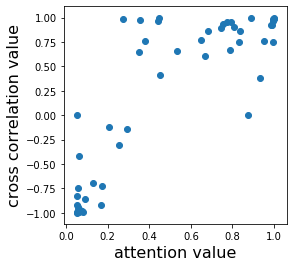"}}
		\centerline{(a)}
	\end{minipage}
	\hfill
	\begin{minipage}{0.24\textwidth}
		\centerline{\includegraphics[width=1\textwidth]{"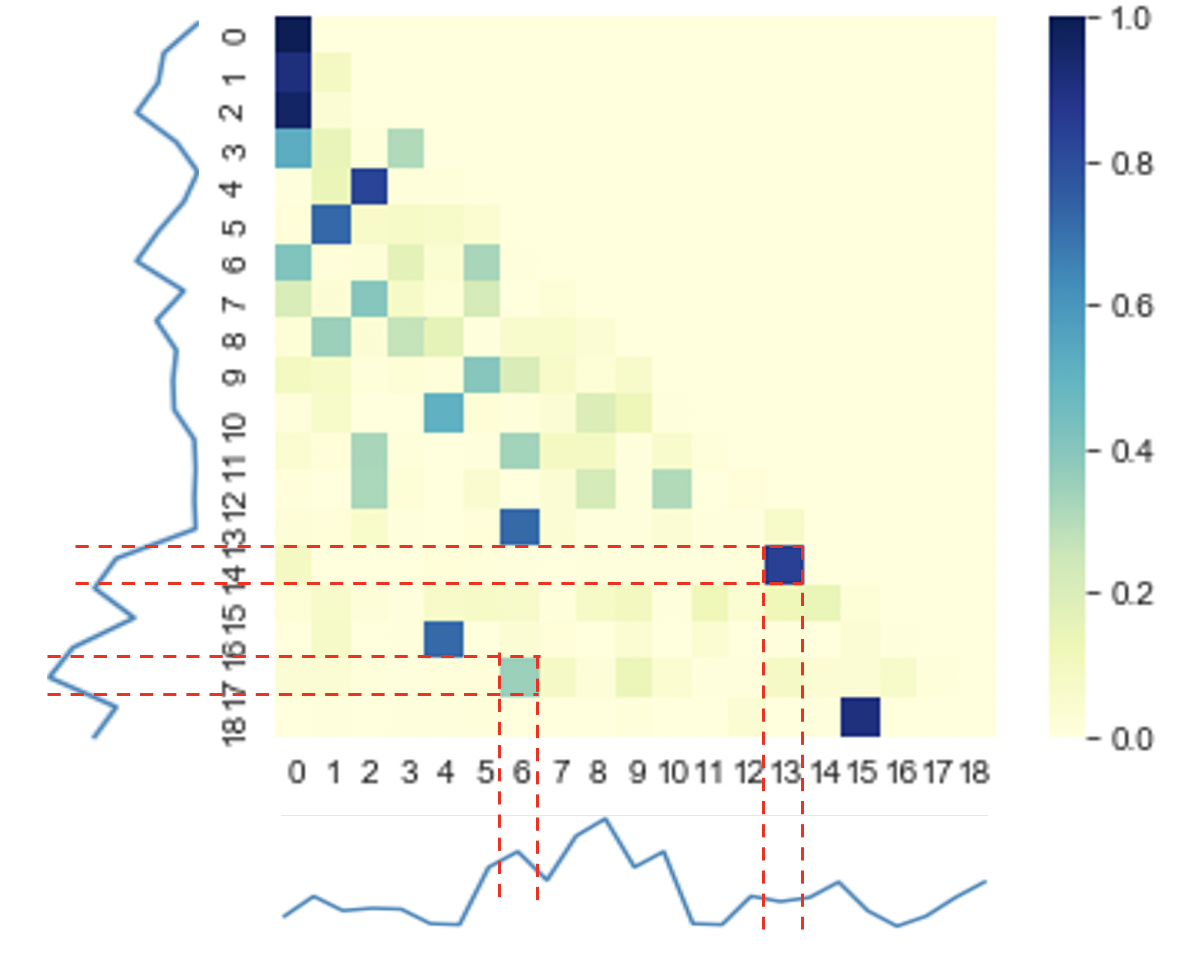"}}
		\centerline{(b)}
	\end{minipage}
	\caption{Case study of the ITA module. (a) Relationship between learned attention weights and cross correlation values for arbitrary GMV pairs in each individual GMV series, (b) Attention  heatmap between a center node and one of its neighbors.}
	\label{fig:atts}
\end{figure}
To better understand the merits of {\model}, we conduct a comprehensive case study for the ITA module, which  intuitively provides convincing evidences for GMV forecasting with attentive weights. 
For intra temporal shift aware attention, we plot the relationship between the learned attention weights and correlation values for arbitrary GMV pairs in each individual GMV series. The negative correlation shown in Fig.~\ref{fig:atts} (a) concludes that similar temporal patterns in single GMV series could be well captured by {\model}. On the other hands, we present an attention heat map between a center node and one of its neighbors in Fig.~\ref{fig:atts} (b) to study the inter temporal shift aware attention. It is not surprising that similar patterns cross two nodes could be find with large attention. It is also worthwhile to note that  a pair of unmatched temporal patterns also attains large attention due to the impact of  auxiliary features.


\section{System Deployment}
\begin{figure}
	\centering
	\includegraphics[width=0.40\textwidth]{"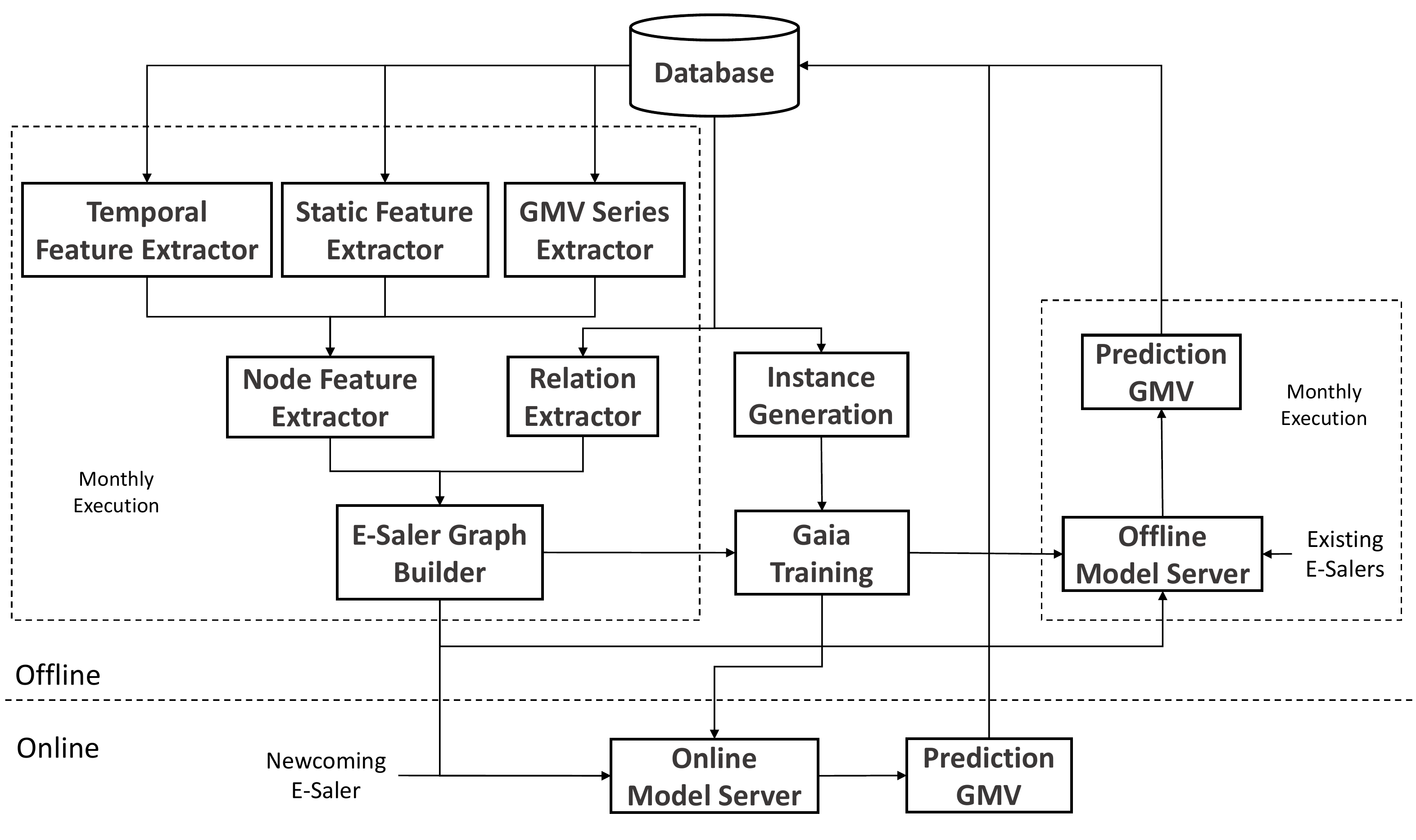"}
	\caption{The deployment details of {\model} in simulated Alipay App environment.}
	\label{fig:deploy}
\end{figure}

To future demonstrate the effectiveness of the proposed {\model}, we deploy it in the Alipay's simulated online environment for GMV forecasting. As shown in Fig.~\ref{fig:deploy}, our deployment follows a hybrid online-offline architecture: offline periodical  training $\rightarrow$ online real-time prediction. 

As mentioned above, a well-established e-seller graph is of crucial importance to support offline training. Assisted by the \emph{Node Feature Extractor} and \emph{Relation Extractor}, abundant features (\ie temporal/static features and GMV series) and relations (\ie same  owner/shareholder and supply chain relationship) are fully explored in an automatic way. It is worthwhile to note that such a simulated pipeline is scheduled monthly to adapt our system to the ever-changing graph structure. 
In the online part,  with regard to a newcoming e-seller, the well-trained {\model} stored in the \emph{Model Server} will make prediction in real time based on its ego-subgraph extracted from the aforementioned e-seller graph. 

Compared to the existed deployed baseline LogTrans, we observe that Gaia achieves 29.1\% improvement on the main metric MAPE (0.117 $\rightarrow$ 0.083).
Our deployed model takes about 10 minutes to predict 2 million e-sellers, the inference time scales linearly with the number of clients.

\section{Conclusion}
In this paper, we propose Gaia, a novel GMV forecasting framework for e-sellers, to address the temporal deficiency and temporal shift issue.
Following the common hierarchical architecture, {\model} jointly models GMV series and auxiliary features in a fine-grained manner with Feature Fusion Layer (FFL), and then learns feature  interaction  along  the  timeline through Temporal Embedding Layer (TEL), followed by the well-designed Inter  and  intra  Temporal  shift  aware  Attention  based Graph Neural Network (ITA-GCN).
Extensive offline and online experiments show the effectiveness of {\model}.


\clearpage
\bibliographystyle{IEEEtran}
\bibliography{ref}
\end{document}